\documentclass[11pt,a4paper]{article}
\usepackage{emnlp2020}
\usepackage{times}
\usepackage{latexsym}
\usepackage{graphicx}
\usepackage{amsmath}
\usepackage{amssymb}
\usepackage{relsize}
\usepackage{dsfont}
\usepackage{xcolor}
\usepackage{multirow}
\usepackage{booktabs}
\usepackage{boldline}
\usepackage{tabularx}
\usepackage{subcaption}
\usepackage{enumitem}
\usepackage{kotex}
\usepackage{footmisc}
\DefineFNsymbols{mySymbols}{{\ensuremath\dagger}{\ensuremath\ddagger}\S\P
   *{**}{\ensuremath{\dagger\dagger}}{\ensuremath{\ddagger\ddagger}}}
\setfnsymbol{mySymbols}
\newcolumntype{Y}{>{\centering\arraybackslash}X}
\newcolumntype{Z}{>{\arraybackslash}X}
\graphicspath{ {./figures/} }
\usepackage{microtype}

\aclfinalcopy 

\title{Multi$^2$OIE: Multilingual Open Information Extraction Based on Multi-Head Attention with BERT}
\author
{
  Youngbin Ro \quad Yukyung Lee \quad Pilsung Kang\thanks{\; Corresponding author} \\
  Korea University, Seoul, Republic of Korea \\
  \texttt{\{youngbin\_ro, yukyung\_lee, pilsung\_kang\}@korea.ac.kr}
}

\begin{document}
\maketitle
\begin{abstract}
In this paper, we propose Multi$^2$OIE, which performs open information extraction (open IE) by combining BERT \citep{devlin-etal-2019-bert} with multi-head attention blocks \citep{10.5555/3295222.3295349}.
Our model is a sequence-labeling system with an efficient and effective argument extraction method.
We use a query, key, and value setting inspired by the Multimodal Transformer \citep{tsai-etal-2019-multimodal} to replace the previously used bidirectional long short-term memory architecture with multi-head attention.
Multi$^2$OIE outperforms existing sequence-labeling systems with high computational efficiency on two benchmark evaluation datasets, Re-OIE2016 and CaRB.
Additionally, we apply the proposed method to multilingual open IE using multilingual BERT.
Experimental results on new benchmark datasets introduced for two languages (Spanish and Portuguese) demonstrate that our model outperforms other multilingual systems without training data for the target languages.
\end{abstract}

\section{Introduction}
Open information extraction (Open IE) \citep{10.5555/1625275.1625705} aims to extract a set of arguments and their corresponding relationship phrases from natural language text.
For example, an open IE system could derive the relational tuple (\emph{was elected}; \emph{The Republican candidate}; \emph{President}) from the given sentence ``\emph{The Republican candidate was elected President.}''
Because the extractions generated by open IE are considered as useful intermediate representations of the source text \citep{10.5555/3061053.3061220}, this method has been applied to various downstream tasks \citep{christensen-etal-2013-towards,ding-etal-2016-knowledge,khot-etal-2017-answering,10.1145/3269206.3271707}.

\begin{figure}[t]
\includegraphics[width=\columnwidth]{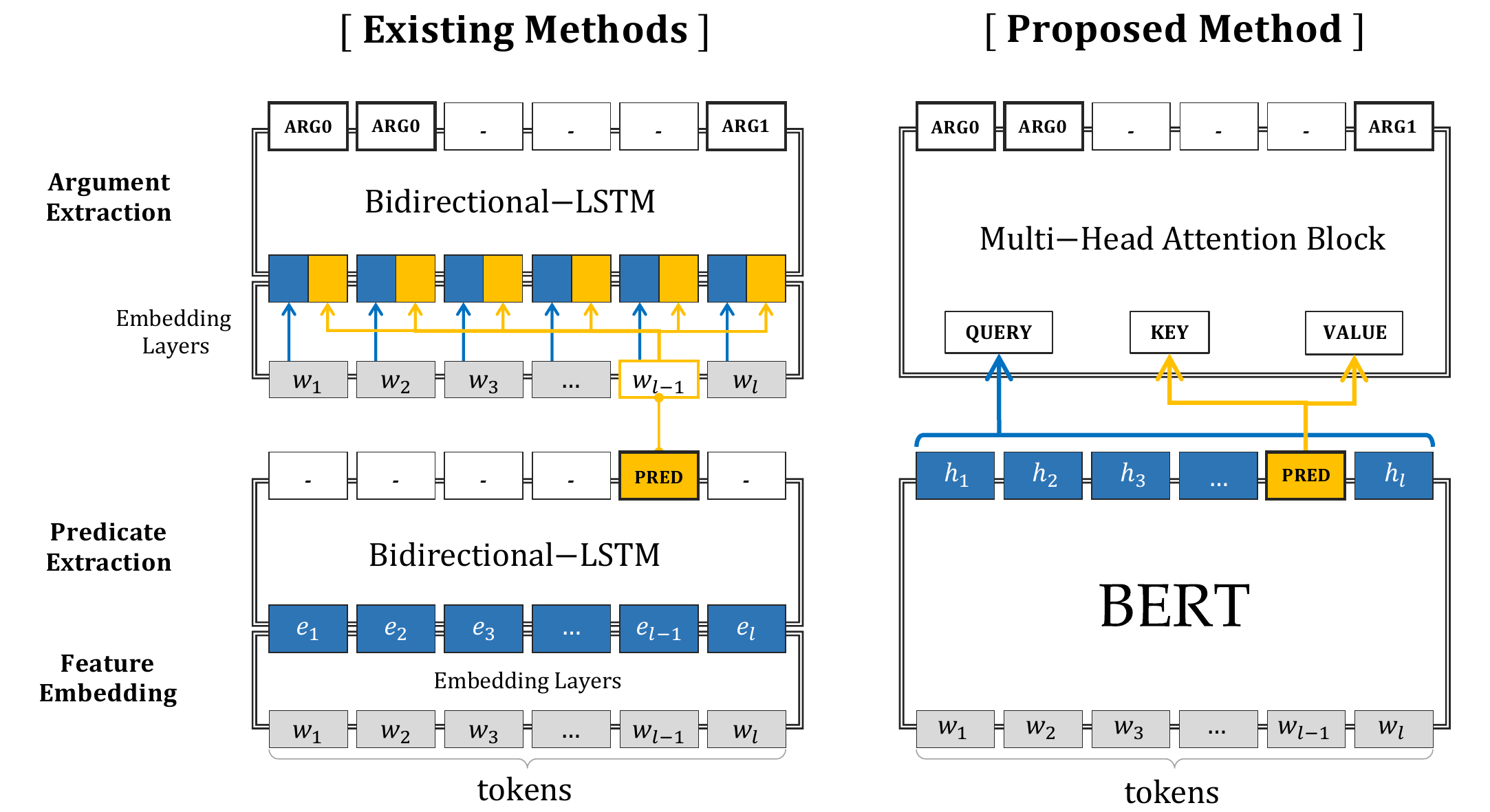}
\caption
{
Comparison between existing extractors and the proposed method.
We use BERT for feature embedding layers and as a predicate extractor.
Predicate information is reflected through multi-head attention instead of simple concatenation.
}
\label{fig:1}
\centering
\end{figure}

Although early open IE systems were largely based on handcrafted features or fine-grained rules \citep{fader-etal-2011-identifying,mausam-etal-2012-open,10.1145/2488388.2488420}, most recent open IE research has focused on deep-neural-network-based supervised learning models.
Such systems are typically based on bidirectional long short-term memory (BiLSTM) and are formulated for two categories: sequence labeling \citep{stanovsky-etal-2018-supervised,Sarhan2019ContextualizedWE,jia2019hybrid} and sequence generation \citep{cui-etal-2018-neural,10.1145/3159652.3159712,bhutani-etal-2019-open}.
The latter enables flexible extraction; however, it is more computationally expensive than the former.
Additionally, generation methods are not suitable for non-English text owing to a lack of training data because they are heavily dependent on in-language supervision \citep{ponti-etal-2019-towards}.
Therefore, we adopted the sequence labeling method to maximize scalability by using (multilingual) BERT \citep{devlin-etal-2019-bert} and multi-head attention \citep{10.5555/3295222.3295349}.
The main advantages of our approach can be summarized as follows:
\begin{itemize}[leftmargin=0.4cm]
\setlength{\itemindent}{0em}
\item
Our model \textbf{\textit{can consider rich semantic and contextual relationships between a predicate and other individual tokens in the same text during sequence labeling by adopting a \underline{multi}-head attention structure}}. Specifically, we apply multi-head attention with the final hidden states from BERT as a query and the hidden states of predicate positions as key-value pairs. This method repeatedly reinforces sentence features by learning attention weights across the predicate and each token \citep{tsai-etal-2019-multimodal}.
Figure \ref{fig:1} presents the difference between the existing sequence labeling methods and the proposed method.
\item
Multi$^2$OIE \textbf{\textit{can operate on \underline{multi}lingual text without non-English training datasets}} by using BERT's multilingual version.
By contrast, for sequence generation systems, performing zero-shot multilingual extraction is much more difficult \citep{ronnqvist-etal-2019-multilingual}.
\item
Our model is more \textbf{\textit{computationally efficient}} than sequence generation systems.
This is because the autoregressive properties of sequence generation create a bottleneck for real-world systems.
This is an important issue for downstream tasks that require processing of large corpora.
\end{itemize}

Experimental results on two English benchmark datasets called Re-OIE2016 \citep{Zhan2019SpanMF} and CaRB \citep{bhardwaj-etal-2019-carb} show that our model yields the best performance among the available sequence-labeling systems.
Additionally, it is demonstrated that the computational efficiency of Multi$^2$OIE is far greater than that of sequence generation systems.
For a multilingual experiment, we introduce multilingual open IE benchmarks (Spanish and Portuguese) constructed by translating and re-annotating the Re-OIE2016 dataset.
Experimental results demonstrate that the proposed Multi$^2$OIE outperforms other multilingual systems without additional training data for non-English languages.
To the best of our knowledge, ours is the first approach using BERT for multilingual open IE\footnote{Although CrossOIE \citep{Cabral2020CrossOIE} considered multilingual BERT in the system, it was not used when extracting the tuples but used only when validating the extracted results.}.
The code and related resources can be found in \url{https://github.com/youngbin-ro/Multi2OIE}.

\section{Background}
\subsection{Multi-Head Attention for Open IE}
In sequence labeling open IE systems, when extracting arguments for a specific predicate, predicate-related features are used as input variables \citep{stanovsky-etal-2018-supervised,Zhan2019SpanMF,jia2019hybrid}.
We analyzed this extraction process from the perspective of multimodal learning \citep{Mangai2010ASO,10.5555/3104482.3104569,10.1109/TPAMI.2018.2798607}, which defines an entire sequence and the corresponding predicate information as a modality.
The most frequently used method for open IE is simple concatenation (Figure \ref{fig:1}, left), which can be interpreted as an early fusion approach.
Simple concatenation has low computational complexity, but requires intensive feature engineering.
It is also highly reliant on the choice of a classifier \citep{doi:10.1142/S1793351X16400158,Liu2018LearnTC}.

Instead, we propose the use of a multi-modality mechanism \citep{tsai-etal-2019-multimodal} to capture the complicated relationships between predicates and other tokens. In our method, multi-head attention is computed by using target modality as a query with source modalities as key-value pairs to adapt the latent information from sources to targets.
This allows our model to assign greater weights to meaningful interactions between modalities.
Accordingly, Multi$^2$OIE uses multi-head attention to reflect predicate information (source modality) throughout a sequence (target modality).
We expect this module to transform a general sentence embedding into a suitable feature for extracting the arguments associated with a specific predicate.

\subsection{Multilingual Open IE}
\label{sec:2.2}
Despite the increasing amount of available web text in languages other than English, most open IE approaches have focused on the English language.
For non-English languages, most systems are heavily reliant on handcrafted features and rules, resulting in limited performance \citep{zhila-gelbukh-2014-open,Oliveira2019DptOIEAP,8903488,GUARASCI2020112954}.
Although some studies have demonstrated the potential of multilingual open IE \citep{faruqui-kumar-2015-multilingual,10.1007/978-3-319-23485-4_72,white-EtAl:2016:EMNLP2016}, most approaches are based on shallow patterns, resulting in low precision \citep{Claro_2019}.

Therefore, we introduce a multilingual-BERT-based open IE system.
BERT provides language-agnostic embedding through its multilingual version and provides excellent zero-shot performance on many classification and labeling tasks \citep{pires-etal-2019-multilingual,wu-dredze-2019-beto,Karthikeyan2020CrossLingualAO}.
In Section \ref{sec:5}, we demonstrate that our multilingual system yields acceptable performance when it is trained using only an English dataset.

\begin{figure*}[ht]
\includegraphics[width=\textwidth]{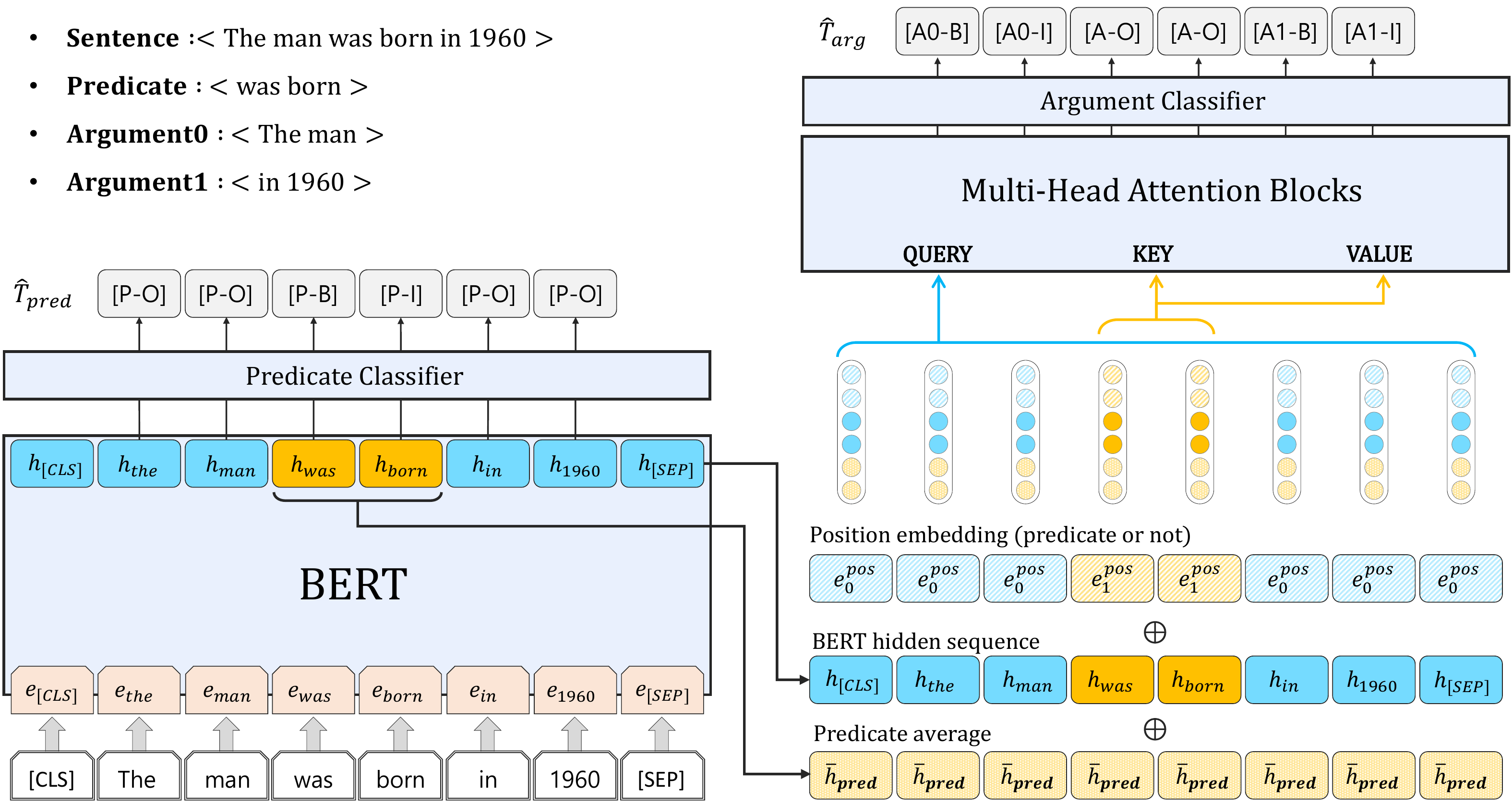}
\caption
{
Architecture of Multi$^2$OIE.
After predicates are extracted using the hidden states of BERT, the hidden sequence, average vector of predicates, and position embedding are concatenated and used as inputs for multi-head attention blocks for argument extraction.
}
\label{fig:2}
\centering
\end{figure*}

\section{Proposed Method}
Multi$^2$OIE extracts relational tuples from a given sentence in two steps. 
The first step is to find all predicates in the sentence.
The second step is to extract the arguments associated with each identified predicate.
The architecture of the proposed model is presented in Figure \ref{fig:2}.

\subsection{Task Formulation}
\label{sec:task_formulation}
Let $S=(w_1, w_2, ..., w_l)$ be an input sentence, where $w_i$ is the $i$-th token and $l$ is the sequence length.
The objective of the proposed model $f$ is to find a set of tags $T=(t_1, t_2, ..., t_l)$, where each element of $T$ indicates one of the “beginning, inside, outside” (BIO) tags \citep{ramshaw-marcus-1995-text}.
However, unlike the method proposed in \citet{stanovsky-etal-2018-supervised}, which uses a predicate head as an input and predicts all tags simultaneously, we first predict a predicate tagset $T_{pred}=(t^p_1, t^p_2, ..., t^p_l)$ using a predicate model $f_{pred}$.
An argument tagset $T_{arg}=(t^a_1, t^a_2, ..., t^a_l)$ is predicted using $f_{arg}$ based on $S$ and $\hat{T}_{pred}$.
Therefore, our model maximizes the following log-likelihood formulation:
\begin{equation} 
\label{eq:1}
\begin{split}
   \mathlarger{\sum}_{i=1}^{l}
   & \Big({\log{p(t^p_i \mid S;\theta_{pred})}} \\
   & + {\log{p(t^a_i \mid \hat{T}_{pred};S;\theta_{pred};\theta_{arg})}}\Big)
\end{split},
\end{equation}
where $\theta_{pred}$ and $\theta_{arg}$ are the trainable parameters of $f_{pred}$ and $f_{arg}$, respectively.
In this formulation, $f_{pred}$ contributes to extracting not only the predicates, but also the arguments.
The loss and gradients derived from argument extraction are also propagated to $\theta_{pred}$ and $\theta_{arg}$.

Additionally, we treat open IE as an $n$-ary extraction task and consider BIO tags for arguments up to ARG3. We refer readers to \citet{stanovsky-etal-2018-supervised} for a more detailed explanation of the BIO sequence labeling policy.

\subsection{Predicate Extraction}
\label{sec:predicate_extraction}
We assume that a given sentence $S$ is tokenized by SentencePiece \citep{kudo-richardson-2018-sentencepiece}.
BERT embeds and encodes $S$ through multiple layers. The final hidden states are defined as $H\in\mathbb{R}^{l \times d}$, where $d$ is the hidden state size of BERT.
$H$ is then fed into a feed-forward network and a softmax layer to calculate the probability that each token is classified into each predicate tag.
The predicted tagset $\hat{T}_{pred}$ is obtained by applying the argmax operation to the softmax outputs.
Finally, the loss for predicate extraction, denoted $L_{pred}$, is calculated as per-token cross-entropy loss.

\subsection{Argument Extraction}
\label{sec:argument_extraction}
A sentence contains one or more predicates.
The argument extraction method described in this section targets only one predicate. The process is simply repeated for multiple predicates.

\paragraph{Input representation}
The inputs for argument extraction are concatenations of the following three features: $H$, $\bar{H}_{pred}$, and $E_{pos}$.
The first feature is the same as the last hidden state of BERT, as discussed in Section \ref{sec:predicate_extraction}.
The second feature is the arithmetic mean vector of hidden states at predicate positions.
We duplicate this vector to match the sequence length $l$ and define it as $\bar{H}_{pred}\in\mathbb{R}^{l \times d}$.
We refer to the true tagset $T_{pred}$ to find the indices of predicates instead of using the predicted tagset $\hat{T}_{pred}$ to achieve more stable training \citep{6795228}.
The final feature $E_{pos}$ is a position embedding of binary values that indicates whether each token is included in the predicate span.
We then concatenate these three features to obtain the input $X\in\mathbb{R}^{l \times d_{mh}}$, where $d_{mh}=2 \cdot d + d_{pos}$ is the dimension of multi-head attention and $d_{pos}$ is the dimension of the position embedding $E_{pos}$.

Following concatenation, $X$ is divided into a query and key-value pairs.
We use $X$ itself as a query, denoted as $X_q$ (target sequence).
Key-value pairs, denoted as $X_k$ and $X_v$ (source sequence), are subsets of $X$ derived from predicate positions.

\begin{figure}[t]
\centering
\includegraphics[scale=0.45]{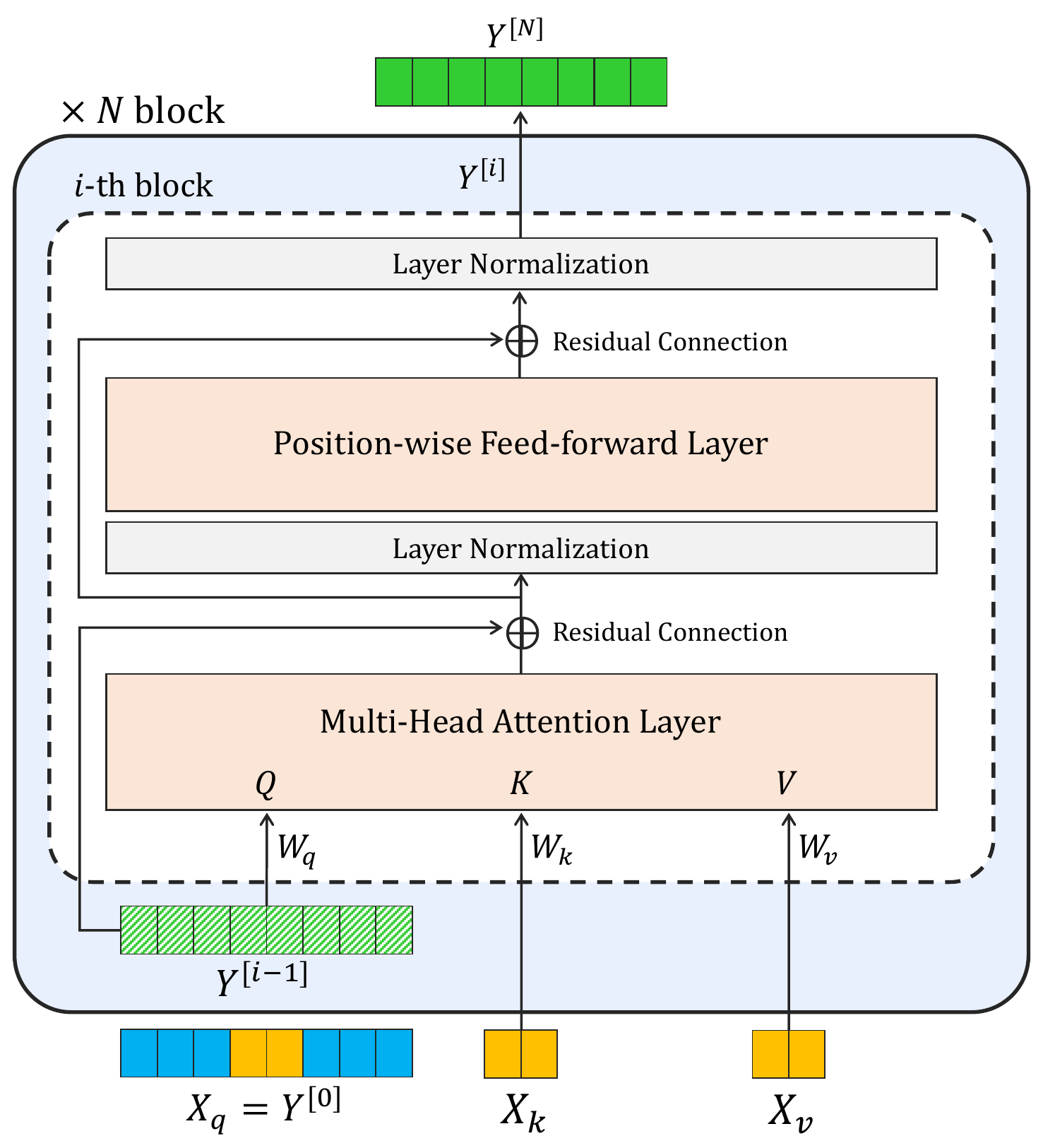}
\caption
{
Multi-head attention blocks for argument extraction.
The architecture consists of $N$ blocks and the output of final block $Y^{[N]}$ is used as the input for the argument classifier.
}
\label{fig:3}
\end{figure}

\paragraph{Multi-head attention block}
The argument extractor consists of $N$ multi-head attention blocks, each of which has a multi-head attention layer followed by a position-wise feed-forward layer, as shown in Figure \ref{fig:3}.

The attention layer is the same as the encoder-decoder attention layer in the original transformer \citep{10.5555/3295222.3295349}.
It first transforms $X_q$, $X_k$, and $X_v$ into $Q=X_qW_q$, $K=X_kW_k$, and $V=X_vW_v$, respectively, where $W_q$, $W_k$, and $W_v$ are weight matrices with dimensions of ($d_{mh} \times d_{mh}$).
Following transformation, the computation of attention is performed for each head as follows:
\begin{equation}
\label{eq:2}
Z_h = \text{Softmax}(\frac{Q_h K^{T}_h}{\sqrt{d_h}})V_h \\.
\end{equation}
Each head is indexed by $h$ and has dimensions of $d_h=\frac{d_{mh}}{n_h}$, where $n_h$ denotes the number of heads.
The attention outputs for each head are then concatenated and linearly transformed.
In addition, we apply residual connections \citep{He2016DeepRL} and layer normalization \citep{Ba2016LayerN} based on the results of prior works on transformers.

The position-wise feed-forward layer consists of two linear transformations surrounding a ReLU activation function.
Residual connections and layer normalization are also applied in this layer.
Finally, the output of the final multi-head attention block is fed into the argument classifier.
The process for obtaining a predicted argument tagset $\hat{T}_{arg}$ and corresponding argument loss $L_{arg}$ is the same as that described in Section \ref{sec:predicate_extraction}.
The final loss for parameter updating is the summation of $L_{pred}$ and $L_{arg}$.

\subsection{Confidence Score}
\label{sec:confidence_score}
In open IE, confidence scores can help control the precision-recall tradeoff of a system.
Multi$^2$OIE provides a confidence score for every extraction by adding the predicate score and all argument scores, as suggested in \citet{Zhan2019SpanMF}.
The score of the predicate and each argument is obtained from the probability value of the \emph{Beginning} tag.
\begin{equation}
\label{eq:5}
CS = p(\text{P-B}) + \mathlarger{\sum}_{i=0}^{3}{\text{ }p(\text{A}_i\text{-B})},
\end{equation}
where the probability values are given by the softmax layer in each extraction step.

\section{Experiments}
\subsection{Experimental Setup}
\label{experimental_setup}
\paragraph{Datasets}
For fair comparisons with other systems, we trained our model using the same dataset used by \citet{Zhan2019SpanMF} \footnote{\url{https://github.com/zhanjunlang/Span_OIE}}.
This dataset was bootstrapped from extractions of the OpenIE4 \citep{10.5555/3061053.3061220}.
For testing data, we used the Re-OIE2016 \citep{Zhan2019SpanMF} and CaRB \citep{bhardwaj-etal-2019-carb}, which were generated via human annotation based on the sentences in the OIE2016 \citep{stanovsky-dagan-2016-creating} dataset.
Table \ref{tab:1} lists the details of the datasets used in this study.

\begin{table}[t]
\centering
\begin{tabular*}{\columnwidth}{cccc} 
\hlineB{3}
\textbf{Split}         & \textbf{Dataset} & \textbf{\# Sents.} & \textbf{\# Tuples} \\ \hlineB{2}
Train                  & OpenIE4          & 1,109,411          & 2,175,294          \\ \hline
\multirow{2}{*}{Dev}   & OIE2016-dev      & 582                & 1,671              \\
                       & CaRB-dev         & 641                & 2,548              \\ \hline
\multirow{2}{*}{Test}  & Re-OIE2016       & 595                & 1,508              \\
                       & CaRB-test        & 641                & 2,715              \\
\hlineB{3}
\end{tabular*}
\caption
{
Numbers of sentences and tuples in each dataset used in this study.
}
\label{tab:1}
\end{table}

\paragraph{Evaluation metrics}
We evaluated each system using the \emph{area under the curve} (AUC) and \emph{F1-score} (F1).
AUC is calculated from a plot of the precision and recall values for all potential cutoffs.
The F1-score is the maximum value among the precision-recall pairs.
We used the evaluation code provided with each test data, which contains the following matching functions: \emph{lexical match}\footnote{\url{https://github.com/gabrielStanovsky/oie-benchmark}} for Re-OIE2016, and \emph{tuple match}\footnote{\url{https://github.com/dair-iitd/CaRB}} for CaRB.
Although the former only considers the existence of words within extractions, the latter is stricter in that it penalizes long extractions \citep{bhardwaj-etal-2019-carb}.

\paragraph{Hyperparameters}
Model hyperparameters were tuned by performing a grid search.
We first trained the model for one epoch with an initial learning rate of 3e-5.
The model contains four multi-head attention blocks with eight attention heads and a 64-dimensional position-embedding layer.
The batch size was set to 128.
The dropout rates for the argument classifier and attention blocks were set to 0.2, respectively.
AdamW \citep{Loshchilov2019DecoupledWD} was used as an optimizer in combination with training heuristics, such as learning rate warmup \citep{Goyal2017AccurateLM} and gradient clipping \citep{10.5555/3042817.3043083}.

\subsection{Baselines}
As baseline models, we selected RnnOIE \citep{stanovsky-etal-2018-supervised}, SpanOIE \citep{Zhan2019SpanMF}, and a few custom systems to evaluate the validity of the multi-head attention blocks (MH).
Although these are all sequence-labeling systems, note that SpanOIE uses the span selection method rather than BIO tagging.
Table \ref{tab:2} presents a summary of the main baselines used in this study.
We also report the results of the following systems developed prior to the use of neural networks: Stanford \citep{angeli-etal-2015-leveraging}, O{\sc llie} \citep{mausam-etal-2012-open}, P{\sc rop}S \citep{Stanovsky2016GettingMO}, ClausIE \citep{10.1145/2488388.2488420}, and OpenIE4.
For these systems, the results were from previous studies \citep{Zhan2019SpanMF,bhardwaj-etal-2019-carb}.

\begin{table}[t]
\centering
\begin{tabular*}{\columnwidth}{lccc} 
\hlineB{3}
                              & \small Method         & $f_{pred}$    & $f_{arg}$          \\ \hlineB{2}
\small BIO                    & \small BIO tagging    & \small BiLSTM & \small BiLSTM      \\
\small BIO+MH                 & \small BIO tagging    & \small BiLSTM & \small MH          \\ \hline
\small SpanOIE                & \small Span selection & \small BiLSTM & \small BiLSTM      \\ 
\small SpanOIE+MH             & \small Span selection & \small BiLSTM & \small MH          \\ \hline
\small BERT+BiLSTM            & \small BIO tagging    & \small BERT   & \small BiLSTM      \\ 
\textbf{\small Multi$^2$OIE} & \small BIO tagging    & \small BERT   & \small MH          \\ \hlineB{3}
\end{tabular*}
\caption
{
Baseline models with difference settings.
}
\label{tab:2}
\end{table}

\begin{figure*}[ht]
\centering
\begin{subfigure}[b]{0.45\textwidth}
\includegraphics[width=\textwidth]{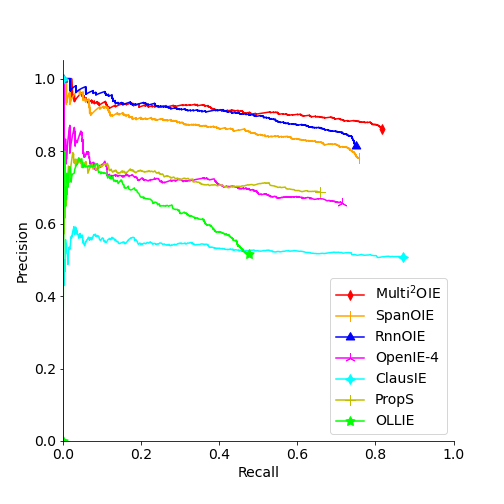}
\caption{Re-OIE2016}
\label{fig:4(a)}
\end{subfigure}
\begin{subfigure}[b]{0.45\textwidth}
\includegraphics[width=\textwidth]{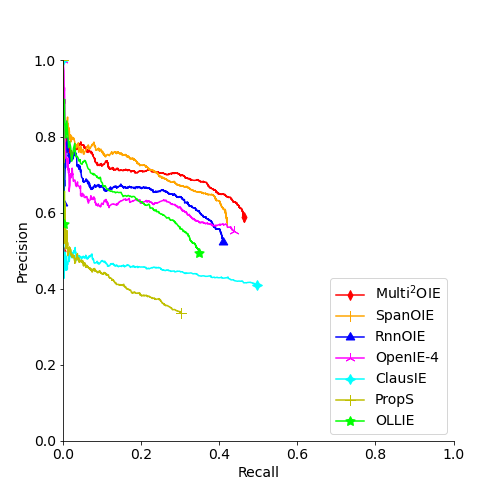}
\caption{CaRB}
\label{fig:4(b)}
\end{subfigure}
\caption
{
Precision-recall curves for each open IE system on two testing datasets. 
}
\label{fig:4}
\end{figure*}

\begin{table*}[ht]
\centering
\begin{tabularx}{0.95\textwidth}{l|YYYY|YYYY}
\hlineB{3}
\multicolumn{1}{c|}{} & \multicolumn{4}{c|}{\textbf{Re-OIE2016}}
                      & \multicolumn{4}{c}{\textbf{CaRB}}                \\ \hline
                      & \multicolumn{1}{c}{\textbf{AUC}}
                      & \multicolumn{1}{c}{\textbf{F1}}
                      & \multicolumn{1}{c}{\small PREC.}
                      & \multicolumn{1}{c|}{\small REC.}
                      & \multicolumn{1}{c}{\textbf{AUC}}
                      & \multicolumn{1}{c}{\textbf{F1}}
                      & \multicolumn{1}{c}{\small PREC.}
                      & \multicolumn{1}{c}{\small REC.}                       \\ \hlineB{2}
Stanford              & 11.5 & 16.7 & - & -    
                      & 13.4 & 23.0 & - & -                              \\
OLLIE                 & 31.3 & 49.5 & - & -     
                      & 22.4 & 41.1 & - & -                              \\
PropS                 & 43.3 & 64.2 & - & -     
                      & 12.6 & 31.9 & - & -                              \\
ClausIE               & 46.4 & 64.2 & - & -   
                      & 22.4 & 44.9 & - & -                              \\
OpenIE4               & 50.9 & 68.3 & - & - 
                      & 27.2 & 48.8 & - & -                              \\ 
RnnOIE                & 68.3 & 78.7 & 84.2 & 73.9
                      & 26.8 & 46.7 & 55.6 & 40.2          \\ \hline
BIO                   & 71.9 & 80.3 & 84.1 & 76.8 
                      & 27.7 & 46.6 & 55.1 & 40.4          \\
BIO+MH                & 71.3 & 81.5 & \textbf{87.0} & 76.6 
                      & 27.3 & 47.5 & 57.2 & 40.7          \\ \hline
SpanOIE               & 65.8 & 77.0 & 79.7 & 74.5 
                      & 30.0 & 49.4 & 60.9 & 41.6          \\
SpanOIE+MH            & 68.0 & 78.8 & 83.1 & 74.9
                      & 30.2 & 50.0 & \textbf{62.2} & 41.8 \\ \hline
BERT+BiLSTM           & 72.1 & 81.3 & 86.0 & 77.0 
                      & 30.6 & 50.6 & 61.3 & 43.1          \\ 
\textbf{Multi$^2$OIE} (ours)    & \textbf{74.6} & \textbf{83.9}
                                & 86.9   & \textbf{81.0}
                                & \textbf{32.6} & \textbf{52.3}
                                & 60.9   & \textbf{45.8}         \\
\hlineB{3}
\end{tabularx}
\caption
{
Performance of Multi$^2$OIE and baseline systems on the Re-OIE2016 and CaRB datasets.
}
\label{tab:3}
\end{table*}

\begin{table*}[ht]
\centering
\begin{tabularx}{0.95\textwidth}{c|Z}
\hlineB{3}
Sentence &  \begin{tabular}{@{}l@{}}
            \emph{At a presentation in the Toronto Pearson International Airport hangar,} \\
            \emph{Celine Dion helped the newly solvent airline debut its new image.}
            \end{tabular} \\ \hline \hline
SpanOIE  &  \begin{tabular}{@{}l@{}}
            \emph{(helped; Celine Dion; the newly solvent airline debut its new image)} \\
            \end{tabular} \\ \hline
Multi$^2$OIE  &   \begin{tabular}{@{}l@{}}
            \emph{(helped; Celine Dion; the newly solvent airline debut its new image;} \\
            \emph{\textbf{At a presentation in the Toronto Pearson International Airport hangar})} \\
            \textbf{\emph{(debut; the newly solvent airline; its new image)}} \\
            \end{tabular} \\
\hlineB{3}
\end{tabularx}
\caption
{
Extraction examples from Multi$^2$OIE and SpanOIE.
The sentences are from the CaRB testing set.
}
\label{tab:4}
\end{table*}

\subsection{Results}
\label{subsection:results}
The performance results for each system on the Re-OIE2016 and CaRB test data are presented in Table \ref{tab:3}.
The precision-recall curves are presented in Figure \ref{fig:4}.
We also present extraction examples from Multi$^2$OIE and SpanOIE in Table \ref{tab:4}.

\paragraph{Overall performance}
Our model outperforms the other systems on all datasets and metrics.
Our model yields average improvements of approximately 6.9\%p and 2.9\%p in terms of F1 for the Re-OIE2016 and CaRB datasets, respectively, compared to the state-of-the-art system (SpanOIE).

Similar to previous studies \citep{stanovsky-etal-2018-supervised,Zhan2019SpanMF}, the excellent performance of Multi$^2$OIE is attributed to improved recall.
As shown in Table \ref{tab:3}, our method achieves the highest recall rate on both datasets.
The examples in Table \ref{tab:4} also demonstrate that our model can extract more tuples from the same sentence.
An additional tuple \emph{(debut; the newly solvent airline; its new image)} is found by Multi$^2$OIE, but not by SpanOIE.
Additionally, Multi$^2$OIE extracts the place information \emph{``At a ... hangar''} for the first tuple, which is omitted by SpanOIE.

\paragraph{Effects of multi-head attention}
We compared three pairs of methods to determine the validity of multi-head attention blocks: (BIO and BIO+MH), (SpanOIE and SpanOIE+MH), and (BERT+BiLSTM and Multi$^2$OIE).
As a result, except for BIO+MH yielding a lower AUC than BIO,
the models with multi-head attention achieve higher performance than the BiLSTM-based models.
This performance improvement is consistent, regardless of the choice of classification method (BIO tagging and span selection).
These results suggest that the use of multi-head attention is superior to simple concatenation in terms of utilizing predicate information.

Additionally, the performance improvement from using MH is greater with BERT than with BiLSTM.
The average performance improvements from BIO to BIO+MH are -0.5\%p (AUC) and 1.1\%p (F1), whereas the improvements from BERT+BiLSTM to Multi$^2$OIE are 2.3\%p (AUC) and 2.2\%p (F1).
This indicates that Multi$^2$OIE has a model architecture that can create synergies between the predicate and argument extractors.

\paragraph{Computational cost}
We measured the training and inference times of each system to evaluate computational efficiency.
As an additional baseline model, we considered a recently published sequence generation system called IMoJIE \citep{kolluru2020imojie}.
It achieved state-of-the-art performance on the CaRB dataset using sequential decoding of tuples conditioned on previous extractions.
For calculating inference times, we selected 641 sentences from the CaRB testing dataset and executed the models on a single TITAN RTX GPU.

Table \ref{tab:5} reveals that Multi$^2$OIE has much greater efficiency than IMoJIE.
Our model only requires 15.5 s to process the 641 sentences, whereas IMoJIE requires more than 3 min, which is a difference of approximately 14 times.
This bottleneck of IMoJIE could be a drawback for downstream tasks, such as knowledge base construction, which must work with large amounts of text.
Considering that the performance difference between the two models is only approximately 1\%p\footnote{IMoJIE achieved (AUC, F1) of (33.3, 53.5) on the CaRB dataset.}, it may be reasonable to use Multi$^2$OIE to process large-scale corpora.
Multi$^2$OIE also exhibits competitive computational costs compared to the other sequence-labeling systems.
Our model has similar training times compared to BERT+BiLSTM, but is faster for inference.
This demonstrates that MH has a positive effect on both efficiency and performance.
In the case of SpanOIE, its span selection method creates bottlenecks for both training and inference.

\begin{table}
\centering
\begin{tabularx}{\columnwidth}{l|YYY}
\hlineB{3}
                       & Training
                       & Inference
                       & Sec./Sent.              \\ \hlineB{2}
\small BERT+BiLSTM     & \textbf{4.5h}
                       & 21.5s
                       & 0.03s                   \\
\small SpanOIE         & 10.2h & 33.8s  & 0.05s  \\
\small IMoJIE          & 7.7h  & 212.2s & 0.33s  \\ \hline
\small \textbf{Multi$^2$OIE} & 4.6h         
                       & \textbf{15.5s}
                       & \textbf{0.02s}          \\ \hlineB{3}
\end{tabularx}
\caption
{
Training and inference times of each system.
}
\label{tab:5}
\end{table}

\section{Multilingual Performance}
\label{sec:5}
As mentioned in Section \ref{sec:2.2}, we trained a multilingual version of Multi$^2$OIE using multilingual BERT and the same training dataset as the English version.
We assumed that data for non-English languages were not available and tested the model's zero-shot performance.
Evaluations were conducted using a dataset generated based on the Re-OIE2016 dataset.

\begin{table}[t]
\centering
\begin{tabularx}{\columnwidth}{l|YYYY}
\hlineB{3}
                  & AUC  & F1   & PREC. & REC. \\ \hlineB{2}
\small EN version & 32.6 & 52.3 & 60.9  & 45.8 \\
\small MT version & 31.5 & 51.9 & 59.5  & 45.9 \\
\hlineB{3}
\end{tabularx}
\caption
{
Comparison between English (EN) and Multilingual (MT) versions of our model on CaRB dataset.
}
\label{tab:6}
\end{table}

\subsection{Experimental setup}
\paragraph{Datasets}
Considering the availability of baseline systems, we selected Spanish and Portuguese as the evaluation dataset languages.
First, all sentences, predicates, and arguments from the Re-OIE2016\footnote{We chose the Re-OIE2016 because the CaRB dataset was originally created not to label sequences but to generate sequences.} dataset were translated into the target languages using Google\footnote{\url{https://cloud.google.com/translate/}}.
To prevent adverse effects from translation errors, we modified the translated sentences to make sure that the back-translated sentences have the same meaning with the original sentence.
After the translation and modification, we manually re-annotated all tuples of the target languages based on the English annotation of Re-OIE2016.

\paragraph{Evaluation metrics}
Because the baseline systems are binary extractors and do not provide confidence scores, we report binary extraction performance without AUC values.
Additionally, although the introduced dataset was generated based on the Re-OIE2016, each system was tested using CaRB's evaluation code for more rigorous evaluation.

\begin{table*}[ht]
\centering
\begin{tabularx}{0.95\textwidth}{c|Z}
\hlineB{3}
Sentence    &  \begin{tabular}{@{}l@{}}
               \small \emph{When the explosion tore through the hut,} \\
               \small \emph{Stauffenberg was convinced that no one in the room could have survived.}
               \end{tabular} \\ \hline \hline
English     &  \begin{tabular}{@{}l@{}}
               \small \emph{(tore; the explosion; through the hut)} \\
               \small \emph{(was convinced; Stauffenberg; that no one in the room could have survived)} \\
               \small \emph{(could have survived; no one in the room)}
               \end{tabular} \\ \hline
Spanish     &  \begin{tabular}{@{}l@{}}
               \small \emph{(desgarró; la explosión; a través de la cabaña)} \\
               \small \emph{(estaba convencido; Stauffenberg; de que nadie en la habitación podría haber sobrevivido)} \\
               \small \emph{(podría haber sobrevivido; nadie en la habitación)}
               \end{tabular} \\ \hline
Portuguese  &  \begin{tabular}{@{}l@{}}
               \small \emph{(rasgou; a explosão; através da cabana)} \\
               \small \emph{(estava convencido; Stauffenberg; de que ninguém na sala poderia ter sobrevivido)} \\
               \small \emph{(poderia ter sobrevivido; ninguém na sala)}
               \end{tabular} \\
\hlineB{3}
\end{tabularx}
\caption
{
Extraction examples from Multi$^2$OIE for each language.
}
\label{tab:7}
\end{table*}

\begin{table}[ht]
\centering
\begin{tabularx}{\columnwidth}{c|l|YYY}
\hlineB{3}
Lang.               & \multicolumn{1}{c|}{System}
                    & F1 & PREC. & REC. \\ \hlineB{2}
\multirow{3}{*}{EN} & \small ArgOE          
                    & 43.4 & 56.6 & 35.2                            \\
                    & \small PredPatt  
                    & 53.1 & 53.9 & 52.3                            \\
                    & \small \textbf{Multi$^2$OIE} 
                    & \textbf{69.3} & \textbf{66.9} & \textbf{71.7} \\ \hline
\multirow{3}{*}{ES} & \small ArgOE  
                    & 39.4 & 48.0 & 33.4                            \\
                    & \small PredPatt  
                    & 44.3 & 44.8 & 43.8                            \\
                    & \small \textbf{Multi$^2$OIE} 
                    & \textbf{60.2} & \textbf{59.1} & \textbf{61.2} \\ \hline
\multirow{3}{*}{PT} & \small ArgOE    
                    & 38.3 & 46.3 & 32.7                   \\
                    & \small PredPatt  
                    & 42.9 & 43.6 & 42.3                            \\
                    & \small \textbf{Multi$^2$OIE} 
                    & \textbf{59.1} & \textbf{56.1} & \textbf{62.5}          \\
\hlineB{3}
\end{tabularx}
\caption
{
Binary extraction performance without confidence scores on the multilingual Re-OIE2016 dataset.
}
\label{tab:8}
\end{table}

\paragraph{Baselines}
Our baseline models were two rule-based multilingual systems: ArgOE \citep{10.1007/978-3-319-23485-4_72} and PredPatt \citep{white-EtAl:2016:EMNLP2016}.
The former takes dependency parses in the CoNLL-X format as inputs.
Similarly, the latter uses language-agnostic patterns of UD structures\footnote{\url{https://universaldependencies.org/}}.

\subsection{Results}
\paragraph{Comparison to the English model}
Prior to comparing the multilingual systems, we evaluated whether Multi$^2$OIE's multilingual version exhibited a satisfactory performance for English compared to the English-only version.
Table \ref{tab:6} lists the performance metrics for the English and multilingual versions of our model on the CaRB dataset.
The performance of the English version was copied from Table \ref{tab:3}.
Although the multilingual version yields lower performance for both metrics compared to the English version, the F1 score is comparable and the recall is higher.
Furthermore, the multilingual version still outperforms the other sequence-labeling systems, indicating that multilingual BERT can successfully construct a Multi$^2$OIE model with favorable performance.

\paragraph{Multilingual performance}
Table \ref{tab:8} lists the performance metrics for each system for the multilingual dataset.
Table \ref{tab:7} contains an example of Multi$^2$OIE's extraction results for each language.
One can see that Multi$^2$OIE outperforms the other systems on all languages.
Similar to the results in Section \ref{subsection:results}, the superiority of our multilingual model is attributed to its high recall.
Multi$^2$OIE yields the highest recall for all languages by approximately 20\%p.
In contrast, ArgOE has relatively high precision, but low recall negatively impacts its F1 score.
PredPatt provides the best balance of precision and recall, but the overall performance is lower than that of our model.

The performance differences between languages are similar for all models.
All models exhibit the best performance for English, followed by Spanish and Portuguese.
Multi$^2$OIE also exhibits performance degradation for non-English languages.
However, considering that our model was never trained to perform open IE tasks on Spanish or Portuguese, its performance is remarkable.
For some non-English sentences, our model extracts the same results as those extracted in the English extraction result, as shown in Table \ref{tab:7}.
This result agrees with the results of previous studies \citep{pires-etal-2019-multilingual,wu-dredze-2019-beto,Karthikeyan2020CrossLingualAO}, which have demonstrated the excellent cross-lingual abilities of multilingual BERT.
Based on these results, we expect that Multi$^2$OIE will also work well on languages other than those considered in this study.

\section{Conclusion}
In this paper, we propose Multi$^2$OIE, which exploits BERT and multi-head attention for the open IE task.
Multi-head attention has the advantage of fusing sentence and predicate features, which adequately reflect predicate information throughout a sentence.
Our model achieved the best performance among sequence labeling models.
Multi$^2$OIE also exhibited superior computational efficiency with competitive performance compared to the state-of-the-art sequence generation systems.
Additionally, a Multi$^2$OIE model trained using multilingual BERT, outperformed the baseline models without training on any non-English languages.

However, some types of extractions, such as nominal relations, conjunctions in arguments, and contextual information, are not considered in Multi$^2$OIE.
Future work could investigate how to apply Multi$^2$OIE to these cases.
For multilingual open IE, performance evaluations and further study on non-alphabetic languages that were not considered in this study can be conducted.

\bibliography{emnlp2020}

\begin{thebibliography}{51}
\expandafter\ifx\csname natexlab\endcsname\relax\def\natexlab#1{#1}\fi

\bibitem[{Angeli et~al.(2015)Angeli, Johnson~Premkumar, and
  Manning}]{angeli-etal-2015-leveraging}
Gabor Angeli, Melvin~Jose Johnson~Premkumar, and Christopher~D. Manning. 2015.
\newblock \href {https://doi.org/10.3115/v1/P15-1034} {Leveraging linguistic
  structure for open domain information extraction}.
\newblock In \emph{Proceedings of the 53rd Annual Meeting of the Association
  for Computational Linguistics and the 7th International Joint Conference on
  Natural Language Processing (Volume 1: Long Papers)}, pages 344--354,
  Beijing, China. Association for Computational Linguistics.

\bibitem[{Ba et~al.(2016)Ba, Kiros, and Hinton}]{Ba2016LayerN}
Jimmy Ba, Jamie~Ryan Kiros, and Geoffrey Hinton. 2016.
\newblock \href {https://arxiv.org/abs/1607.06450} {Layer normalization}.
\newblock arXiv:1607.06450.

\bibitem[{Baltrusaitis et~al.(2019)Baltrusaitis, Ahuja, and
  Morency}]{10.1109/TPAMI.2018.2798607}
Tadas Baltrusaitis, Chaitanya Ahuja, and Louis-Philippe Morency. 2019.
\newblock \href {https://doi.org/10.1109/TPAMI.2018.2798607} {Multimodal
  machine learning: A survey and taxonomy}.
\newblock \emph{IEEE Transactions on Pattern Analysis and Machine
  Intelligence}, 41(2):423–443.

\bibitem[{Banko et~al.(2007)Banko, Cafarella, Soderland, Broadhead, and
  Etzioni}]{10.5555/1625275.1625705}
Michele Banko, Michael~John Cafarella, Stephen Soderland, Matt Broadhead, and
  Oren Etzioni. 2007.
\newblock \href {https://dl.acm.org/doi/10.5555/1625275.1625705} {Open
  information extraction from the web}.
\newblock In \emph{Proceedings of the 20th International Joint Conference on
  Artifical Intelligence}, IJCAI’07, page 2670–2676, San Francisco, CA,
  USA.

\bibitem[{Bhardwaj et~al.(2019)Bhardwaj, Aggarwal, and
  Mausam}]{bhardwaj-etal-2019-carb}
Sangnie Bhardwaj, Samarth Aggarwal, and Mausam Mausam. 2019.
\newblock \href {https://doi.org/10.18653/v1/D19-1651} {{C}a{RB}: A
  crowdsourced benchmark for open {IE}}.
\newblock In \emph{Proceedings of the 2019 Conference on Empirical Methods in
  Natural Language Processing and the 9th International Joint Conference on
  Natural Language Processing (EMNLP-IJCNLP)}, pages 6262--6267, Hong Kong,
  China. Association for Computational Linguistics.

\bibitem[{Bhutani et~al.(2019)Bhutani, Suhara, Tan, Halevy, and
  Jagadish}]{bhutani-etal-2019-open}
Nikita Bhutani, Yoshihiko Suhara, Wang-Chiew Tan, Alon Halevy, and
  Hosagrahar~Visvesvaraya Jagadish. 2019.
\newblock \href {https://doi.org/10.18653/v1/N19-1239} {Open information
  extraction from question-answer pairs}.
\newblock In \emph{Proceedings of the 2019 Conference of the North {A}merican
  Chapter of the Association for Computational Linguistics: Human Language
  Technologies, Volume 1 (Long and Short Papers)}, pages 2294--2305,
  Minneapolis, Minnesota. Association for Computational Linguistics.

\bibitem[{Cabral et~al.(2020)Cabral, Glauber, Souza, and
  Claro}]{Cabral2020CrossOIE}
Bruno Cabral, Rafael Glauber, Marlo Souza, and Daniela Claro. 2020.
\newblock \href {https://doi.org/10.1007/978-3-030-41505-1_35} {Crossoie:
  Cross-lingual classifier for open information extraction}.
\newblock In \emph{Computational Processing of the Portuguese Language}, pages
  368--378.

\bibitem[{Christensen et~al.(2013)Christensen, {Mausam}, Soderland, and
  Etzioni}]{christensen-etal-2013-towards}
Janara Christensen, {Mausam}, Stephen Soderland, and Oren Etzioni. 2013.
\newblock \href {https://www.aclweb.org/anthology/N13-1136} {Towards coherent
  multi-document summarization}.
\newblock In \emph{Proceedings of the 2013 Conference of the North {A}merican
  Chapter of the Association for Computational Linguistics: Human Language
  Technologies}, pages 1163--1173, Atlanta, Georgia. Association for
  Computational Linguistics.

\bibitem[{Claro et~al.(2019)Claro, Souza, Xavier, and Oliveira}]{Claro_2019}
Daniela~Barreiro Claro, Marlo Souza, Clarissa~Castellã Xavier, and Leandro
  Oliveira. 2019.
\newblock \href {https://doi.org/10.3390/info10070228} {Multilingual open
  information extraction: Challenges and opportunities}.
\newblock \emph{Information}, 10(7):228.

\bibitem[{Cui et~al.(2018)Cui, Wei, and Zhou}]{cui-etal-2018-neural}
Lei Cui, Furu Wei, and Ming Zhou. 2018.
\newblock \href {https://doi.org/10.18653/v1/P18-2065} {Neural open information
  extraction}.
\newblock In \emph{Proceedings of the 56th Annual Meeting of the Association
  for Computational Linguistics (Volume 2: Short Papers)}, pages 407--413,
  Melbourne, Australia. Association for Computational Linguistics.

\bibitem[{Del~Corro and Gemulla(2013)}]{10.1145/2488388.2488420}
Luciano Del~Corro and Rainer Gemulla. 2013.
\newblock \href {https://doi.org/10.1145/2488388.2488420} {Clausie:
  Clause-based open information extraction}.
\newblock In \emph{Proceedings of the 22nd International Conference on World
  Wide Web}, WWW ’13, page 355–366, New York, NY, USA. Association for
  Computing Machinery.

\bibitem[{Devlin et~al.(2019)Devlin, Chang, Lee, and
  Toutanova}]{devlin-etal-2019-bert}
Jacob Devlin, Ming-Wei Chang, Kenton Lee, and Kristina Toutanova. 2019.
\newblock \href {https://doi.org/10.18653/v1/N19-1423} {{BERT}: Pre-training of
  deep bidirectional transformers for language understanding}.
\newblock In \emph{Proceedings of the 2019 Conference of the North {A}merican
  Chapter of the Association for Computational Linguistics: Human Language
  Technologies, Volume 1 (Long and Short Papers)}, pages 4171--4186,
  Minneapolis, Minnesota. Association for Computational Linguistics.

\bibitem[{Ding et~al.(2016)Ding, Zhang, Liu, and
  Duan}]{ding-etal-2016-knowledge}
Xiao Ding, Yue Zhang, Ting Liu, and Junwen Duan. 2016.
\newblock \href {https://www.aclweb.org/anthology/C16-1201} {Knowledge-driven
  event embedding for stock prediction}.
\newblock In \emph{Proceedings of {COLING} 2016, the 26th International
  Conference on Computational Linguistics: Technical Papers}, pages 2133--2142.

\bibitem[{Ergun et~al.(2016)Ergun, Akyuz, Sert, and
  Liu}]{doi:10.1142/S1793351X16400158}
Hilal Ergun, Yusuf~Caglar Akyuz, Mustafa Sert, and Jianquan Liu. 2016.
\newblock \href {https://doi.org/10.1142/S1793351X16400158} {Early and late
  level fusion of deep convolutional neural networks for visual concept
  recognition}.
\newblock \emph{International Journal of Semantic Computing}, 10(03):379--397.

\bibitem[{Fader et~al.(2011)Fader, Soderland, and
  Etzioni}]{fader-etal-2011-identifying}
Anthony Fader, Stephen Soderland, and Oren Etzioni. 2011.
\newblock \href {https://www.aclweb.org/anthology/D11-1142} {Identifying
  relations for open information extraction}.
\newblock In \emph{Proceedings of the 2011 Conference on Empirical Methods in
  Natural Language Processing}, pages 1535--1545, Edinburgh, Scotland, UK.
  Association for Computational Linguistics.

\bibitem[{Faruqui and Kumar(2015)}]{faruqui-kumar-2015-multilingual}
Manaal Faruqui and Shankar Kumar. 2015.
\newblock \href {https://doi.org/10.3115/v1/N15-1151} {Multilingual open
  relation extraction using cross-lingual projection}.
\newblock In \emph{Proceedings of the 2015 Conference of the North {A}merican
  Chapter of the Association for Computational Linguistics: Human Language
  Technologies}, pages 1351--1356, Denver, Colorado. Association for
  Computational Linguistics.

\bibitem[{Gamallo and Garcia(2015)}]{10.1007/978-3-319-23485-4_72}
Pablo Gamallo and Marcos Garcia. 2015.
\newblock \href {https://doi.org/10.1007/978-3-319-23485-4_72} {Multilingual
  open information extraction}.
\newblock In \emph{Progress in Artificial Intelligence (EPIA 2015)}, pages
  711--722.

\bibitem[{Goyal et~al.(2017)Goyal, Doll{\'a}r, Girshick, Noordhuis, Wesolowski,
  Kyrola, Tulloch, Jia, and He}]{Goyal2017AccurateLM}
Priya Goyal, Piotr Doll{\'a}r, Ross Girshick, Pieter Noordhuis, Lukasz
  Wesolowski, Aapo Kyrola, Andrew Tulloch, Yangqing Jia, and Kaiming He. 2017.
\newblock \href {https://arxiv.org/abs/1706.02677} {Accurate, large minibatch
  sgd: Training imagenet in 1 hour}.
\newblock arXiv:1706.02677.

\bibitem[{Guarasci et~al.(2020)Guarasci, Damiano, Minutolo, Esposito, and
  Pietro}]{GUARASCI2020112954}
Raffaele Guarasci, Emanuele Damiano, Aniello Minutolo, Massimo Esposito, and
  Giuseppe~De Pietro. 2020.
\newblock \href {https://doi.org/https://doi.org/10.1016/j.eswa.2019.112954}
  {Lexicon-grammar based open information extraction from natural language
  sentences in italian}.
\newblock \emph{Expert Systems with Applications}, 143:112954.

\bibitem[{He et~al.(2016)He, Zhang, Ren, and Sun}]{He2016DeepRL}
Kaiming He, Xiangyu Zhang, Shaoqing Ren, and Jian Sun. 2016.
\newblock \href {https://ieeexplore.ieee.org/document/7780459} {Deep residual
  learning for image recognition}.
\newblock \emph{2016 IEEE Conference on Computer Vision and Pattern Recognition
  (CVPR)}, pages 770--778.

\bibitem[{Jia and Xiang(2019)}]{jia2019hybrid}
Shengbin Jia and Yang Xiang. 2019.
\newblock \href {https://arxiv.org/abs/1908.01761} {Hybrid neural tagging model
  for open relation extraction}.
\newblock arXiv:1908.01761.

\bibitem[{Karthikeyan et~al.(2020)Karthikeyan, Wang, Mayhew, and
  Roth}]{Karthikeyan2020CrossLingualAO}
Kaliyaperumal Karthikeyan, Zihan Wang, Stephen Mayhew, and Dan Roth. 2020.
\newblock \href {https://cogcomp.seas.upenn.edu/papers/KWMR20.pdf}
  {Cross-lingual ability of multilingual bert: An empirical study}.
\newblock In \emph{Proceedings of the International Conference on Learning
  Representations (ICLR)}.

\bibitem[{Khot et~al.(2017)Khot, Sabharwal, and
  Clark}]{khot-etal-2017-answering}
Tushar Khot, Ashish Sabharwal, and Peter Clark. 2017.
\newblock \href {https://doi.org/10.18653/v1/P17-2049} {Answering complex
  questions using open information extraction}.
\newblock In \emph{Proceedings of the 55th Annual Meeting of the Association
  for Computational Linguistics (Volume 2: Short Papers)}, pages 311--316,
  Vancouver, Canada. Association for Computational Linguistics.

\bibitem[{Kolluru et~al.(2020)Kolluru, Aggarwal, Rathore, Mausam, and
  Chakrabarti}]{kolluru2020imojie}
Keshav Kolluru, Samarth Aggarwal, Vipul Rathore, Mausam Mausam, and Soumen
  Chakrabarti. 2020.
\newblock \href {https://www.aclweb.org/anthology/2020.acl-main.521/} {Imojie:
  Iterative memory-based joint open information extraction}.
\newblock In \emph{The 58th Annual Meeting of the Association for Computational
  Linguistics (ACL)}, Seattle, U.S.A. Association for Computational
  Linguistics.

\bibitem[{Kudo and Richardson(2018)}]{kudo-richardson-2018-sentencepiece}
Taku Kudo and John Richardson. 2018.
\newblock \href {https://doi.org/10.18653/v1/D18-2012} {{S}entence{P}iece: A
  simple and language independent subword tokenizer and detokenizer for neural
  text processing}.
\newblock In \emph{Proceedings of the 2018 Conference on Empirical Methods in
  Natural Language Processing: System Demonstrations}, pages 66--71, Brussels,
  Belgium. Association for Computational Linguistics.

\bibitem[{Liu et~al.(2018)Liu, Li, Xu, and Natarajan}]{Liu2018LearnTC}
Kuan Liu, Yanen Li, Ning Xu, and Premkumar Natarajan. 2018.
\newblock \href {https://arxiv.org/abs/1805.11730} {Learn to combine modalities
  in multimodal deep learning}.
\newblock arXiv:1805.11730.

\bibitem[{Loshchilov and Hutter(2019)}]{Loshchilov2019DecoupledWD}
Ilya Loshchilov and Frank Hutter. 2019.
\newblock \href {https://openreview.net/forum?id=Bkg6RiCqY7} {Decoupled weight
  decay regularization}.
\newblock In \emph{Proceedings of the International Conference on Learning
  Representations (ICLR)}.

\bibitem[{Mangai et~al.(2010)Mangai, Samanta, Das, and
  Chowdhury}]{Mangai2010ASO}
Utthara~Gosa Mangai, Suranjana Samanta, Sukhendu Das, and Pinaki~Roy Chowdhury.
  2010.
\newblock \href {https://www.tandfonline.com/doi/abs/10.4103/0256-4602.64604}
  {A survey of decision fusion and feature fusion strategies for pattern
  classification}.
\newblock \emph{Iete Technical Review}, 27:293--307.

\bibitem[{Mausam(2016)}]{10.5555/3061053.3061220}
Mausam. 2016.
\newblock \href {https://www.ijcai.org/Proceedings/16/Papers/604.pdf} {Open
  information extraction systems and downstream applications}.
\newblock In \emph{Proceedings of the Twenty-Fifth International Joint
  Conference on Artificial Intelligence}, IJCAI’16, page 4074–4077.

\bibitem[{{Mausam} et~al.(2012){Mausam}, Schmitz, Soderland, Bart, and
  Etzioni}]{mausam-etal-2012-open}
{Mausam}, Michael Schmitz, Stephen Soderland, Robert Bart, and Oren Etzioni.
  2012.
\newblock \href {https://www.aclweb.org/anthology/D12-1048} {Open language
  learning for information extraction}.
\newblock In \emph{Proceedings of the 2012 Joint Conference on Empirical
  Methods in Natural Language Processing and Computational Natural Language
  Learning}, pages 523--534, Jeju Island, Korea. Association for Computational
  Linguistics.

\bibitem[{Ngiam et~al.(2011)Ngiam, Khosla, Kim, Nam, Lee, and
  Ng}]{10.5555/3104482.3104569}
Jiquan Ngiam, Aditya Khosla, Mingyu Kim, Juhan Nam, Honglak Lee, and Andrew Ng.
  2011.
\newblock \href {http://www.icml-2011.org/papers/399_icmlpaper.pdf} {Multimodal
  deep learning}.
\newblock In \emph{Proceedings of the 28th International Conference on
  International Conference on Machine Learning}, ICML’11, page 689–696,
  Madison, WI, USA.

\bibitem[{de~Oliveira and Claro(2019)}]{Oliveira2019DptOIEAP}
Leandro~Souza de~Oliveira and Daniela~Barreiro Claro. 2019.
\newblock \href {http://repositorio.ufba.br/ri/handle/ri/30719} {Dptoie: a
  portuguese open information extraction system based on dependency analysis}.
\newblock \emph{Computer Speech and Language}, under review.

\bibitem[{Pascanu et~al.(2013)Pascanu, Mikolov, and
  Bengio}]{10.5555/3042817.3043083}
Razvan Pascanu, Tomas Mikolov, and Yoshua Bengio. 2013.
\newblock \href {http://proceedings.mlr.press/v28/pascanu13.html} {On the
  difficulty of training recurrent neural networks}.
\newblock In \emph{Proceedings of the 30th International Conference on
  International Conference on Machine Learning - Volume 28}, ICML’13, page
  III–1310–III–1318.

\bibitem[{Pires et~al.(2019)Pires, Schlinger, and
  Garrette}]{pires-etal-2019-multilingual}
Telmo Pires, Eva Schlinger, and Dan Garrette. 2019.
\newblock \href {https://doi.org/10.18653/v1/P19-1493} {How multilingual is
  multilingual {BERT}?}
\newblock In \emph{Proceedings of the 57th Annual Meeting of the Association
  for Computational Linguistics}, pages 4996--5001, Florence, Italy.
  Association for Computational Linguistics.

\bibitem[{Ponti et~al.(2019)Ponti, Vuli{\'c}, Cotterell, Reichart, and
  Korhonen}]{ponti-etal-2019-towards}
Edoardo~Maria Ponti, Ivan Vuli{\'c}, Ryan Cotterell, Roi Reichart, and Anna
  Korhonen. 2019.
\newblock \href {https://doi.org/10.18653/v1/D19-1288} {Towards zero-shot
  language modeling}.
\newblock In \emph{Proceedings of the 2019 Conference on Empirical Methods in
  Natural Language Processing and the 9th International Joint Conference on
  Natural Language Processing (EMNLP-IJCNLP)}, pages 2900--2910, Hong Kong,
  China. Association for Computational Linguistics.

\bibitem[{Ramshaw and Marcus(1995)}]{ramshaw-marcus-1995-text}
Lance Ramshaw and Mitchell Marcus. 1995.
\newblock \href {https://www.aclweb.org/anthology/W95-0107} {Text chunking
  using transformation-based learning}.
\newblock In \emph{Proceedings of the Third ACL Workshop on Very Large
  Corpora}, pages 82--94.

\bibitem[{R{\"o}nnqvist et~al.(2019)R{\"o}nnqvist, Kanerva, Salakoski, and
  Ginter}]{ronnqvist-etal-2019-multilingual}
Samuel R{\"o}nnqvist, Jenna Kanerva, Tapio Salakoski, and Filip Ginter. 2019.
\newblock \href {https://www.aclweb.org/anthology/W19-6204} {Is multilingual
  {BERT} fluent in language generation?}
\newblock In \emph{Proceedings of the First NLPL Workshop on Deep Learning for
  Natural Language Processing}, pages 29--36, Turku, Finland.

\bibitem[{Sarhan and Spruit(2019)}]{Sarhan2019ContextualizedWE}
Injy Sarhan and Marco Spruit. 2019.
\newblock \href {https://doi.org/10.1007/978-3-030-23281-8_31} {Contextualized
  word embeddings in a neural open information extraction model}.
\newblock In \emph{Natural Language Processing and Information Systems}, pages
  359--367.

\bibitem[{Stanovsky and Dagan(2016)}]{stanovsky-dagan-2016-creating}
Gabriel Stanovsky and Ido Dagan. 2016.
\newblock \href {https://doi.org/10.18653/v1/D16-1252} {Creating a large
  benchmark for open information extraction}.
\newblock In \emph{Proceedings of the 2016 Conference on Empirical Methods in
  Natural Language Processing}, pages 2300--2305, Austin, Texas. Association
  for Computational Linguistics.

\bibitem[{Stanovsky et~al.(2016)Stanovsky, Ficler, Dagan, and
  Goldberg}]{Stanovsky2016GettingMO}
Gabriel Stanovsky, Jessica Ficler, Ido Dagan, and Yoav Goldberg. 2016.
\newblock \href {https://arxiv.org/abs/1603.01648} {Getting more out of syntax
  with props}.
\newblock arXiv:1603.01648.

\bibitem[{Stanovsky et~al.(2018)Stanovsky, Michael, Zettlemoyer, and
  Dagan}]{stanovsky-etal-2018-supervised}
Gabriel Stanovsky, Julian Michael, Luke Zettlemoyer, and Ido Dagan. 2018.
\newblock \href {https://doi.org/10.18653/v1/N18-1081} {Supervised open
  information extraction}.
\newblock In \emph{Proceedings of the 2018 Conference of the North {A}merican
  Chapter of the Association for Computational Linguistics: Human Language
  Technologies, Volume 1 (Long Papers)}, pages 885--895, New Orleans,
  Louisiana. Association for Computational Linguistics.

\bibitem[{Sun et~al.(2018)Sun, Li, Wang, Fan, Feng, and
  Li}]{10.1145/3159652.3159712}
Mingming Sun, Xu~Li, Xin Wang, Miao Fan, Yue Feng, and Ping Li. 2018.
\newblock \href {https://doi.org/10.1145/3159652.3159712} {Logician: A unified
  end-to-end neural approach for open-domain information extraction}.
\newblock In \emph{Proceedings of the Eleventh ACM International Conference on
  Web Search and Data Mining}, WSDM ’18, page 556–564, New York, NY, USA.
  Association for Computing Machinery.

\bibitem[{Tsai et~al.(2019)Tsai, Bai, Liang, Kolter, Morency, and
  Salakhutdinov}]{tsai-etal-2019-multimodal}
Yao-Hung~Hubert Tsai, Shaojie Bai, Paul~Pu Liang, J.~Zico Kolter,
  Louis-Philippe Morency, and Ruslan Salakhutdinov. 2019.
\newblock \href {https://doi.org/10.18653/v1/P19-1656} {Multimodal transformer
  for unaligned multimodal language sequences}.
\newblock In \emph{Proceedings of the 57th Annual Meeting of the Association
  for Computational Linguistics}, pages 6558--6569, Florence, Italy.
  Association for Computational Linguistics.

\bibitem[{Vaswani et~al.(2017)Vaswani, Shazeer, Parmar, Uszkoreit, Jones,
  Gomez, Kaiser, and Polosukhin}]{10.5555/3295222.3295349}
Ashish Vaswani, Noam Shazeer, Niki Parmar, Jakob Uszkoreit, Llion Jones, Aidan
  Gomez, \L~ukasz Kaiser, and Illia Polosukhin. 2017.
\newblock \href
  {http://papers.nips.cc/paper/7181-attention-is-all-you-need.pdf} {Attention
  is all you need}.
\newblock In \emph{Advances in Neural Information Processing Systems}, pages
  6000--6010.

\bibitem[{Wang et~al.(2019)Wang, He, and Zhou}]{8903488}
Chengyu Wang, Xiaofeng He, and Aoying Zhou. 2019.
\newblock \href {https://doi.org/10.1109/TKDE.2019.2953839} {Open relation
  extraction for chinese noun phrases}.
\newblock \emph{IEEE Transactions on Knowledge and Data Engineering}, PP:1--1.

\bibitem[{White et~al.(2016)White, Reisinger, Sakaguchi, Vieira, Zhang,
  Rudinger, Rawlins, and {Van Durme}}]{white-EtAl:2016:EMNLP2016}
Aaron~Steven White, Drew Reisinger, Keisuke Sakaguchi, Tim Vieira, Sheng Zhang,
  Rachel Rudinger, Kyle Rawlins, and Benjamin {Van Durme}. 2016.
\newblock \href {https://aclweb.org/anthology/D16-1177} {{Universal
  Decompositional Semantics on Universal Dependencies}}.
\newblock In \emph{Proceedings of the 2016 Conference on Empirical Methods in
  Natural Language Processing}, pages 1713--1723, Austin, Texas. Association
  for Computational Linguistics.

\bibitem[{Williams and Zipser(1989)}]{6795228}
Ronald Williams and David Zipser. 1989.
\newblock \href {https://doi.org/10.1162/neco.1989.1.2.270} {A learning
  algorithm for continually running fully recurrent neural networks}.
\newblock \emph{Neural Computation}, 1(2):270--280.

\bibitem[{Wu and Dredze(2019)}]{wu-dredze-2019-beto}
Shijie Wu and Mark Dredze. 2019.
\newblock \href {https://doi.org/10.18653/v1/D19-1077} {Beto, bentz, becas: The
  surprising cross-lingual effectiveness of {BERT}}.
\newblock In \emph{Proceedings of the 2019 Conference on Empirical Methods in
  Natural Language Processing and the 9th International Joint Conference on
  Natural Language Processing (EMNLP-IJCNLP)}, pages 833--844, Hong Kong,
  China. Association for Computational Linguistics.

\bibitem[{Wu et~al.(2018)Wu, Wu, Kao, and Yin}]{10.1145/3269206.3271707}
Tien-Hsuan Wu, Zhiyong Wu, Ben Kao, and Pengcheng Yin. 2018.
\newblock \href {https://doi.org/10.1145/3269206.3271707} {Towards practical
  open knowledge base canonicalization}.
\newblock In \emph{Proceedings of the 27th ACM International Conference on
  Information and Knowledge Management}, CIKM ’18, page 883–892, New York,
  NY, USA. Association for Computing Machinery.

\bibitem[{Zhan and Zhao(2020)}]{Zhan2019SpanMF}
Junlang Zhan and Hai Zhao. 2020.
\newblock \href {https://doi.org/10.1609/aaai.v34i05.6497} {Span model for open
  information extraction on accurate corpus}.
\newblock \emph{Proceedings of the AAAI Conference on Artificial Intelligence},
  34:9523--9530.

\bibitem[{Zhila and Gelbukh(2014)}]{zhila-gelbukh-2014-open}
Alisa Zhila and Alexander Gelbukh. 2014.
\newblock \href {https://doi.org/10.3115/v1/P14-3011} {Open information
  extraction for {S}panish language based on syntactic constraints}.
\newblock In \emph{Proceedings of the {ACL} 2014 Student Research Workshop},
  pages 78--85, Baltimore, Maryland, USA. Association for Computational
  Linguistics.

\end{thebibliography}
\bibliographystyle{acl_natbib}

\end{document}